\documentclass[conference]{IEEEtran}
\IEEEoverridecommandlockouts
\usepackage{graphicx}
\usepackage{multirow}
\usepackage{caption}
\usepackage{subcaption}
\usepackage{cite}
\usepackage{comment}
\usepackage{amsmath}
\usepackage{amssymb}
\usepackage{adjustbox}
\usepackage{booktabs}
\usepackage{amsfonts}
\usepackage{array}
\usepackage{amsmath}
\usepackage{amssymb}
\usepackage{algorithm}
\usepackage{algorithmic}
\usepackage{lmodern,babel,adjustbox,booktabs,multirow}
\usepackage{float}
\usepackage{placeins}
\usepackage{comment}
\usepackage{xcolor}
\usepackage{array}
\usepackage{textcomp}
\usepackage{caption}
\usepackage{subcaption}
\usepackage{enumitem}
\usepackage{etoolbox}
\usepackage[utf8]{inputenc}
\apptocmd{\thebibliography}{\small}{}{}
\patchcmd{\thebibliography}{\leftmargin\labelwidth}{\leftmargin\labelwidth\itemsep=0pt\parsep=0pt\topsep=0pt}{}{}

\begin{document}

\title{FGA: Fourier-Guided Attention Network for Crowd Count Estimation \\
}

\author{\IEEEauthorblockN{Yashwardhan Chaudhuri}
\IEEEauthorblockA{
 \textit{IIIT-Delhi}\\
yashwardhan20417@iiitd.ac.in}
 \and
 \IEEEauthorblockN{Ankit Kumar}
 \IEEEauthorblockA{
 \textit{IIT-Bombay}\\
 ak670676@gmail.com}
 \and
\IEEEauthorblockN{Arun Balaji Buduru}
 \IEEEauthorblockA{
 \textit{IIIT-Delhi}\\
 arunb@iiitd.ac.in}
 \and
 \IEEEauthorblockN{Adel Alshamrani}
 \IEEEauthorblockA{
 \textit{University of Jeddah}\\
 asalshamrani@uj.edu.sa}
 }

\maketitle

\begin{abstract}

Crowd counting is gaining societal relevance, particularly in domains of Urban Planning, Crowd Management, and Public Safety. This paper introduces Fourier-guided attention (FGA), a novel attention mechanism for crowd count estimation designed to address the inefficient full-scale global pattern capture in existing works on convolution-based attention networks. FGA efficiently captures multi-scale information, including full-scale global patterns, by utilizing Fast-Fourier Transformations (FFT) along with spatial attention for global features and convolutions with channel-wise attention for semi-global and local features. The architecture of FGA involves a dual-path approach: (1) a path for processing full-scale global features through FFT, allowing for efficient extraction of information in the frequency domain, and (2) a path for processing remaining feature maps for semi-global and local features using traditional convolutions and channel-wise attention. This dual-path architecture enables FGA to seamlessly integrate frequency and spatial information, enhancing its ability to capture diverse crowd patterns. We apply FGA in the last layers of two popular crowd-counting works, CSRNet and CANNet, to evaluate the module's performance on benchmark datasets such as ShanghaiTech-A, ShanghaiTech-B, UCF-CC-50, and JHU++ crowd. The experiments demonstrate a notable improvement across all datasets based on Mean-Squared-Error (MSE) and Mean-Absolute-Error (MAE) metrics, showing comparable performance to recent state-of-the-art methods. Additionally, we illustrate the interpretability using qualitative analysis, leveraging Grad-CAM heatmaps, to show the effectiveness of FGA in capturing crowd patterns.
\end{abstract}

\begin{IEEEkeywords}
Crowd Count Estimation, Fast Fourier Transformation, Attention, Channel Attention, Spatial Attention, CNN
\end{IEEEkeywords}

\section{Introduction}
\label{sec:intro}

Crowd count estimation (or crowd counting) involves estimating the number of individuals in a particular crowd scene. Its utility in public safety, crowd management, urban planning, and healthcare makes it relevant in computer vision. Crowd counting becomes challenging in large crowd scenes where problems such as large foreground-background imbalances, occlusions, and perspective distortion arise frequently. Density-based crowd-counting methods\cite{gao2019scar}\cite{gao2020counting}\cite{zhang2016single}\cite{li2018csrnet} is widely accepted as a solution to this problem. It involves predicting crowd density maps (Figure 1), where the sum of pixel values in a density map gives the number of people in it. Crowd counting has progressed significantly since the introduction of deep learning, with works such as MCNN\cite{zhang2016single},  CSRNet\cite{li2018csrnet}, OURS-CAN\cite{liu2019context},  ASPDNet\cite{gao2020counting} being widely accepted as a possible solution.
\begin{figure}[!t]
  \centering
  \includegraphics[width=1\columnwidth ,height=0.16\textheight]{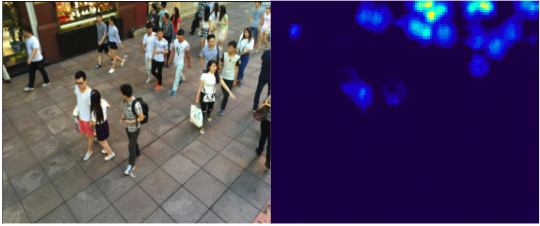}
\caption{\textbf{Left:} Image of a crowd scene as input to the neural network.  \textbf{Right:} Density map of The crowd scene. Brighter spots are noticed in the top right corner of the density map where crowd density is higher and becomes less visible towards the left where there Is less crowd density.}
  \label{fig1}
\end{figure}
\begin{figure*}[!t]
  \centering
  \includegraphics[width=0.9\textwidth, height=0.32\textheight]{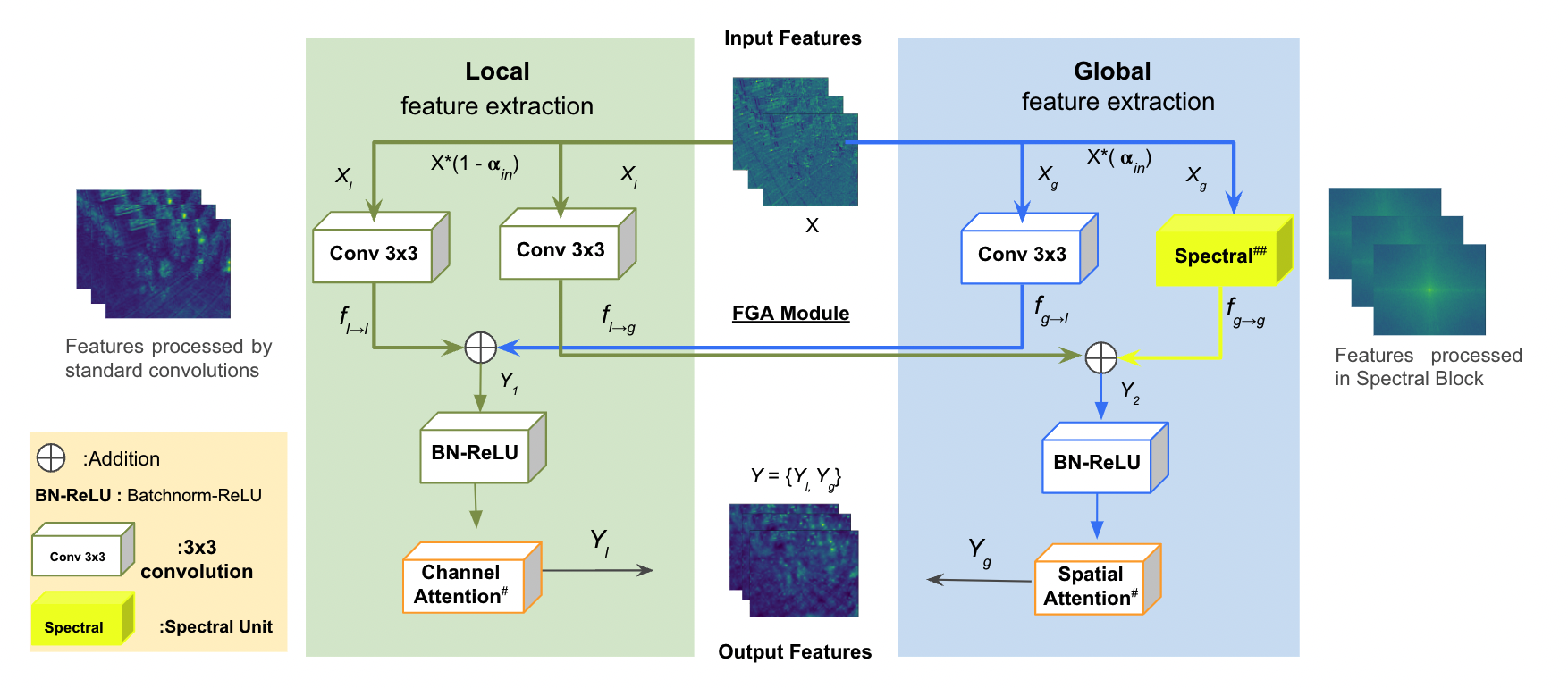}
  \caption{\textbf{FGA Module:} The module has two feature extraction sections, as shown in the above figure. \textcolor{black}{\textbf{Left:}} The local feature extraction takes a fraction of the input feature maps for local processing. \textcolor{black}{\textbf{Right:}} The global feature extraction takes another fraction as input from feature maps for global processing. The spectral block captures full-scale global features. \textcolor{black}{\textbf{\#\#}}: Refer to Figure 3 for more details. \textcolor{black}{\textbf{\#}}: refer to Figure 4 for more details on attention blocks.}
  \label{fig2}
\end{figure*}

Most density map-based solutions employ a CNN\cite{lecun1995convolutional}-based approach to regress crowd counts. The following methods are promising but remain constrained by the convolutional kernel's receptive field, leading to inefficient capture of global or long-range patterns in models. A global receptive field in crowd counting allows the network to capture information from a wider context, ensuring that it considers the overall layout of the crowd and accounts for variations in crowd density distribution.
Many attention-based works such as MAN\cite{lin2022boosting}, DMCNet\cite{wang2023dynamic}, JANet\cite{he2022jointly}, DA2Net\cite{zhai20232} have gained popularity because of their ability to understand large-scale dependencies through attention networks. BBA-net \cite{hou2020bba}, proposes an attention-based network designed to capture the fine-grained details in spatial locations. RFSNet \cite{zhang2023recurrent} introduces a patch-wise recurrent self-attention network for spatio-temporal crowd counting in video.
Although these CNN methods perform exceptionally well, they only utilize convolutions capable of processing information in a local neighbourhood, ignoring the large-scale pixel-to-pixel context; thus, using convolutional layers alone can prove inefficient for understanding full-scale global patterns.
To solve this problem, we suggest a novel neural architecture named the Fourier-Guided Attention (FGA) module, drawing inspiration from the Fast Fourier Convolution (FFC) \cite{chi2020fast} and \cite{gao2019scar}. This module is designed for long-range context-aware crowd counting, seamlessly combining FFC, spatial attention, and channel attention into a single unit. FFC operates in the spatial and frequency domains, allowing FGA to process information in both local and global receptive fields. At the same time, different attention mechanisms focus on amplifying the fine-grained features from different parts of the input sequence to capture relevant information. The proposed framework can be integrated into existing crowd-counting methods to attend to full-scale global features.

To conclude, our contributions are mainly threefold:
\begin{itemize}
  
  \item A novel dual-path architecture utilizing FFC and Attention mechanisms incorporating local and global context into a single pluggable unit for existing works.

  \item Two simple FGA integrated architectures with comparable performances to the state-of-the-art methods in crowd counting.
 
  \item A thorough qualitative and quantitative evaluation of our proposed approach with four public benchmark datasets: Shanghai-Tech Part A, Shanghai-Tech Part B, JHU++ crowd, and UCF-CC-50, along with an extensive ablation study to understand the contribution of each part in the FGA module.

\end{itemize}



\section{Related Work}
\subsection{Density Based estimation}
Most recent crowd-counting methods employ a variation of CNN to predict density maps from crowd images. The density maps represent the spatial distribution of people in the image, with brighter regions depicting higher crowd density. Numerous approaches have been proposed to mitigate the impact of scale variations, such as MCNN \cite{zhang2016single}, which employs a multi-column architecture with different kernel sizes to capture scale variation across a crowd scene effectively. CSRNet \cite{li2018csrnet} and CANNET \cite{liu2019context} use single-column with a dilated convolution layer. certain encoder-decoder methods like MRCNet \cite{bahmanyar2019mrcnet}, \cite{jiang2019crowd} incorporate contextual and detailed local information by integrating high- and low-level features through several lateral connections.
The point-based framework \cite{song2021rethinking} utilizes one-to-one matching with doted annotation, while \cite{song2021choose} makes patch-level feature selection.
Some methods \cite{wan2020modeling}\cite{wang2020distribution} \cite{sindagi2020learning} reduce the Gaussian noise during annotation and density map generation to give better predictions. In GauNet\cite{cheng2022rethinking}, the convolution filter is replaced by locally connected Gaussian kernels. The work proposes a low-rank approximation accompanied by translation invariance to implement Gaussian convolution for density estimation.
 
Although efficient in capturing local features of crowd scenes, the methods are inefficient in capturing large-scale pixel-by-pixel information. Attention-based models are introduced to address the mentioned issue.

\subsection{Attention based methods}
Attention mechanisms \cite{woo2018cbam} \cite{hu2018squeeze} have proven to be highly efficient in addressing various computer vision challenges. Spatial attention focuses on learning a weighting map that emphasizes specific spatial coordinates within the feature map and channel attention is responsible for learning a weighting map that highlights important feature channels within the feature map. Notable works such as \cite{gao2019scar} \cite{jiang2020attention} \cite{guo2019dadnet} \cite{varior2019multi} \cite{gao2020counting} integrate the spatial and channel attention. SCAR \cite{gao2019scar}  Uses novel spatial and channel attention and attempts to extract more discriminative features among different channels, which helps the model identify heads in an image

The RANet \cite{zhang2019relational} advocates the self-attention mechanism by accounting for both short and long-range interdependence of pixels and implementing it using local self-attention (LSA) and global self-attention (GSA) along with a relational module. \cite{hossain2019crowd} propose a Scale-Aware Attention Network(SANet) for local and global features
\cite{zhang2019attentional} 
Introduce conditional random fields (CRFs) to aggregate
multi-scale features within the encoder-decoder network. It incorporates a non-local attention mechanism implemented as inter- and intra-layer attentions to expand the receptive field, capturing long-range dependencies to overcome huge scale variations.
\cite{jiang2020attention} purpose an attention-scaling network ASNet and DSNet to learn auto-scaling for density estimation. In \cite{gao2020counting}, spatial and channel attention along with the spatial pyramidal pooling and deformation convolution for object counting. ADCrowdNet, proposed by \cite{liu2019adcrowdnet}, employs an attention map generator to identify regions of interest and estimate congestion levels for improved density map estimation in crowd counting. Although these works have good performance and attending to large-scale patterns, are unable to capture purely global features due to the limited kernel size of convolutions. To this end, we propose FGA Module, a simple plug-in module for existing CNN-based works that can capture full-scale global information using the frequency domain for understanding global patterns in a crowd.
\section{Method}
\label{sec:methodology}

\subsection{Model}
Our work introduces the FGA module for extracting full-scale global, semi-global and local patterns as shown in figure 2. 
Given input features \(X \in \mathbb{R}^{H \times W \times C}\), where \(H \times W\) represents the spatial resolution and \(C\) represents the number of channels, separating factor \(\alpha_{\text{in}}\) determines the fraction of the input feature map \(X \in \mathbb{R}^{H \times W \times C}\) used for global feature extraction $X_g$. The remaining features are used for the local extraction $X_l$. The parameter for separating factor ranges, denoted as  \(\alpha_{\text{in}}\), is within the interval [0, 1]. 
\begin{equation}
\mathbf{ X_g \in \mathbb{R}^{H \times W \times (\alpha_{\text{in}})C}}
\end{equation}
\begin{equation}
\mathbf{X_l \in \mathbb{R}^{H \times W \times (1 - \alpha_{\text{in}})C}}
\end{equation}
$X_l$ passes through two separate 3x3 convolutional layers to produce $f_{l\to l}$ and $f_{l\to g}$, used for processing local and semi-global context. $X_g$  is processed independently in a 3x3 convolutional layer and a spectral block to produce $f_{g\to l}$ and $f_{g\to g}$, used for processing semi-global and global feature extraction. The spectral block utilizes\cite{katznelson2004introduction} to update a signal value in the spectral domain, thereby influencing the entire feature map.
We introduce semi-global information into the two extraction blocks by adding $f_{l\to g}$ to $f_{g\to g}$ and $f_{g\to l}$ to $f_{l\to l}$, as shown in the equation.
\begin{equation}
    Y_1 = f_{g\to l} + f_{l\to l}
\end{equation}
\begin{equation}
    Y_2 = f_{l\to g} + f_{g\to g}
\end{equation}
We pass $Y_1$ and $Y_2$ through the batch-normalization ReLU combination. Finally, $Y_1$ and $Y_2$ are passed through Channel, spatial attention\cite{gao2019scar} respectively. The spatial attention amplifies the global features for each feature map in $Y_2$ to get $Y_g$, whereas the channel attention amplifies relevant information across channels in $Y_1$ to get $Y_l$.
The final output $Y$ is combined set $\{ Y_g, Y_l \}$.

\begin{figure}[!t]
  \centering
  \includegraphics[width=0.4\columnwidth]{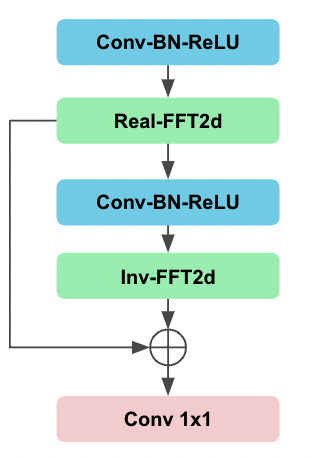}
  \caption{\textbf{Spectral Block:} The image above explains the functioning of the spectral block. \textbf{Conv-BN-ReLU:} refers to convolution, batch normalization, ReLU combination. \textbf{Real-FFT2d:} refers to the fast-Fourier transformation of real features. \textbf{Inv-FFT2d:} Refers to Fourier domain to real domain transformation. \textbf{Conv 1x1 :} refers to 1x1 size convolutions.}
  \label{fig:Spectral}
\end{figure}
\subsection {Spectral Block}
The spectral block employs the Cooley-Tukey algorithm \cite{cooley1965algorithm} for efficient Fast Fourier Transform (FFT) of input feature maps. The spectral block converts original spatial features into the frequency domain, also known as the spectral domain. Subsequently, the spectral data is globally updated within the frequency domain and transformed back into the spatial domain.
Applying the 2-D FFT to the input \(X\), the complex frequency domain \(X[u, v]\)  is represented as
\begin{equation}
   X_g[u, v] = \sum_{n=0}^{M-1} \sum_{m=0}^{N-1} x[m, n] \cdot e^{-j2\pi\left(\frac{un}{M} + \frac{vm}{N}\right)}
\end{equation}
where:
   \(X[u, v]\) is the complex coefficient at frequency index \((u, v)\) in the frequency domain.
   \(x[m, n]\) is the pixel value at spatial position \((m, n)\) in the image.
   \(M\) is the height of the image.
   \(N\) is the width of the image.
   \(u\) and \(v\) are the frequency indices ranging from 0 to \(M-1\) and 0 to \(N-1\), respectively.
We vertically stack the imaginary parts and real parts as independent features to simplify the computation of FFT-processed features.  
If \(c\) is the number of channels in \(X\), we obtain a tensor \(Y\) with twice the number of channels is given as;

\begin{equation}
Y[c, u, v] = \begin{cases} 
                  X_r[u, v] & \text{if } c < C \text{ (Real part)} \\
                  X_i[u, v] & \text{if } C \leq c < 2C \text{ (Imaginary part)} \\
               \end{cases}
\end{equation}
The resultant tensor is treated as completely computable, following practices \cite{chi2019fast} applying Batch Normalization (BN), Rectified Linear Unit (ReLU) and Convolution operation in the frequency domain to extract relation between different pixels, Which gives the tensor  \(Z\) as 
\begin{equation}  
Z[c, u, v] = \text{ReLU}(\text{BN}(\textbf{Conv}(Y[c, u, v])))
\end{equation}
In the final step, the results of tensor Z are transformed back into complex numbers by splitting them along the auxiliary dimension, separating the real and the imaginary parts. The inverse 2-D FFT operation applied on tensor Z to produce an output tensor with real values is given as follows.

\begin{equation}
\begin{split}
X_{\text{out}}[m, n] = \frac{1}{MN} \sum_{u=0}^{M-1} \sum_{v=0}^{N-1} e^{j2\pi\left(\frac{un}{M}\right. 
+ \left. \frac{vm}{N}\right)} \cdot \\ (Z_r[u, v] + jZ_i[u, v])
\end{split}
\end{equation}
This equation calculates the inverse 2-D IFFT by summing the contributions from both the real and imaginary parts for each pixel \((n, m)\). The result \(X_{\text{out}}[m, n]\) represents the spatial domain representation of the image after applying the inverse transform. We further utilize spatial and channel attention to enhance local and global features.
\begin{figure}[!t]
  \centering
  \includegraphics[width=0.6\columnwidth]{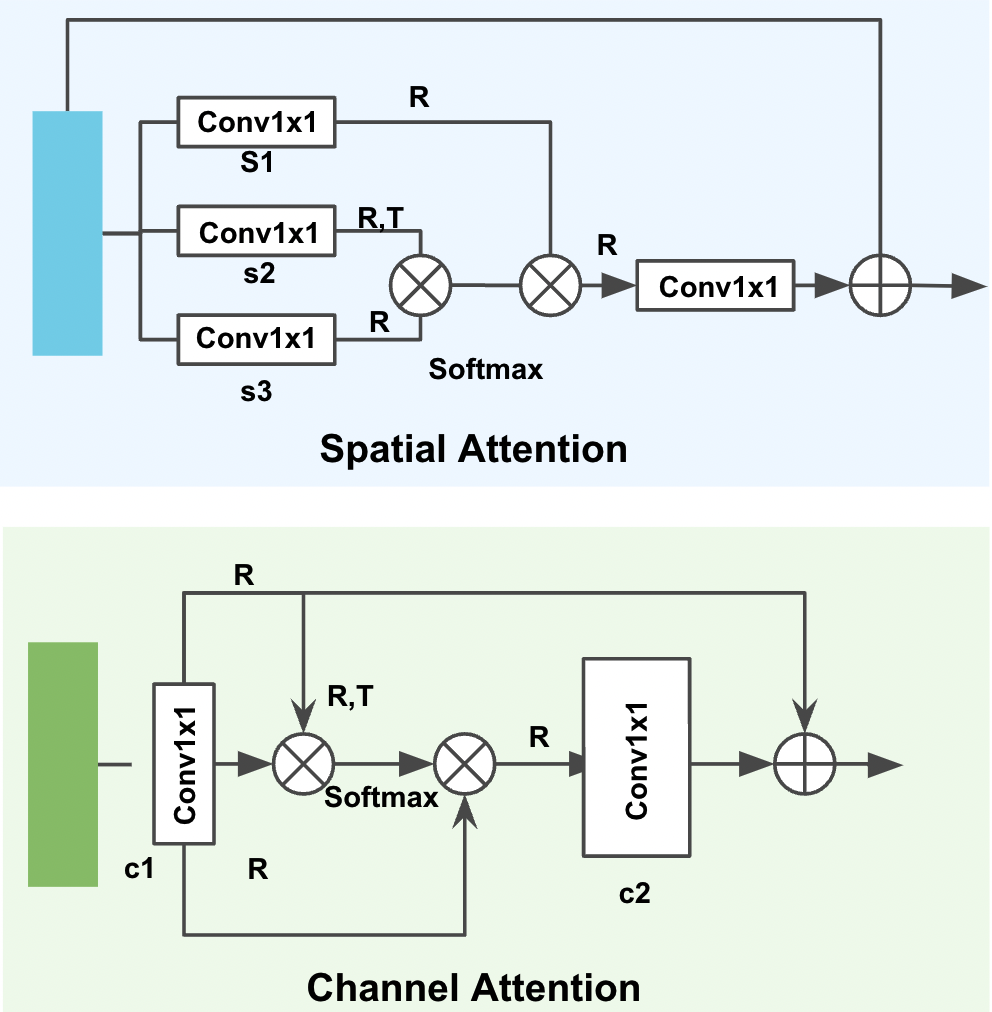}
    \caption{Attention Blocks: The image above shows spatial attention used in the global extraction block in Figure 2. and the channel attention block in the local feature extraction block. R: Resize T: Transpose}
  \label{fig:Attnetion}
\end{figure}
\begin{figure*}[htbp]
    \centering
    \begin{subfigure}[b]{0.33\textwidth}
        \centering
        \includegraphics[width=\textwidth]{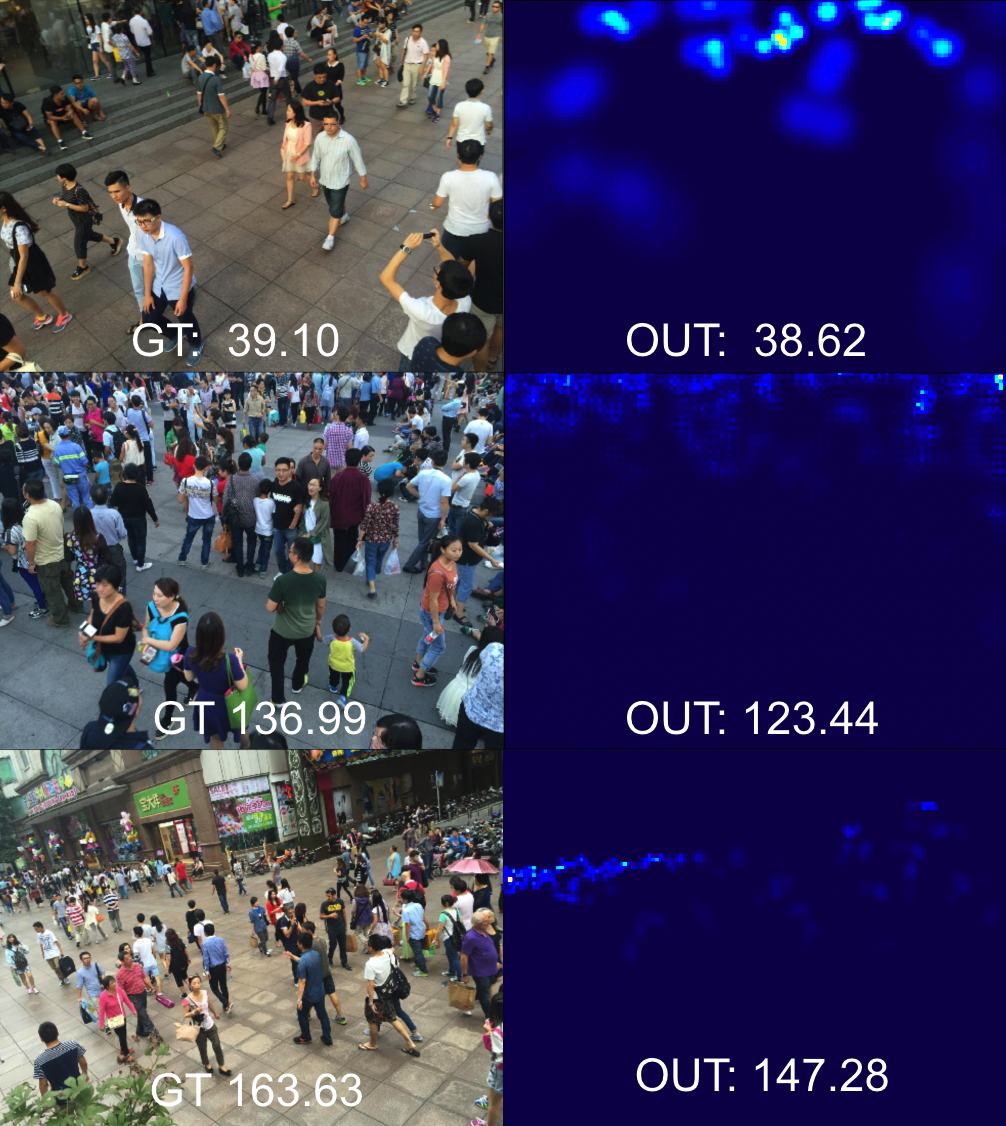}
        \caption{\textbf{FGA-CSRNet}}
        \label{fig:sub1}
    \end{subfigure}%
    \begin{subfigure}[b]{0.34\textwidth}
        \centering
        \includegraphics[width=\textwidth , height =0.284\textheight]{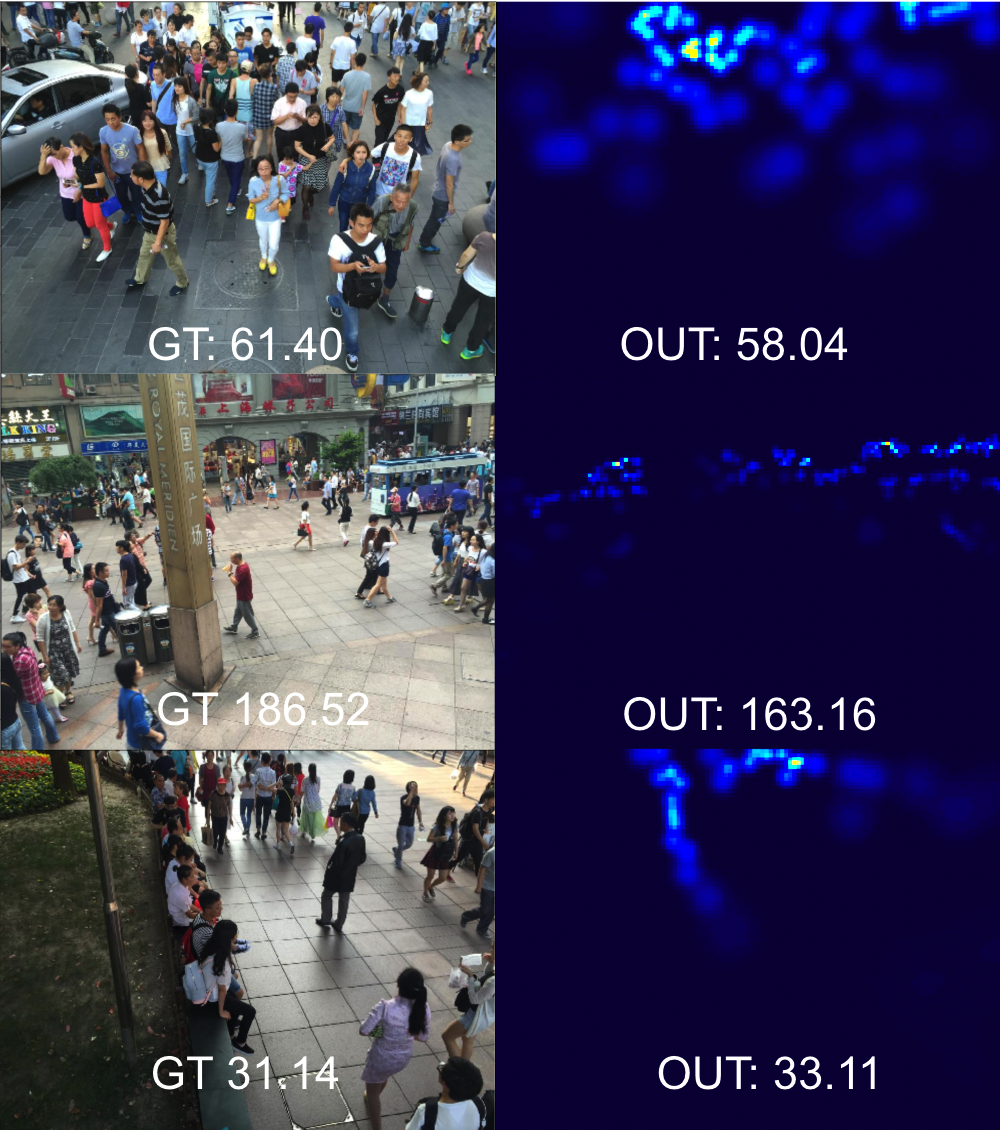}
        \caption{\textbf{FGA-CANNet}}
        \label{fig:sub2}
    \end{subfigure}

    \caption{\textbf{Counting samples from varying crowd distributions in ShanghaiTech-B combination:} Each output density map is shown right adjacent to the input image when given to two different models with FGA module.}
    \label{fig:visual_samples}
\end{figure*}
\subsection{Attention Blocks}
We adopt attention mechanisms \cite{hu2018squeeze} \cite{woo2018cbam} \cite{gao2019scar} to enhance the features; spatial attention allows the model to focus on specific regions of crowd variation across a particular feature map, whereas channel attention is related to overall crowd characteristics across channels.
For Spatial Attention, we transform the input feature maps, $X \in \mathbb{R}^{C \times H \times W}$, into three feature spaces $S_1$, $S_2$, and $S_3$by using $1 \times 1$ convolutional layers. The corresponding weight matrix of features is given as $ S_1(X) = W_{S1}(x)$, \quad $S_2(X) = W_{S2}(x)$, \quad $S_3(X) = W_{S3}(x)$.  The Transpose operations are then performed on $S_1(X)$, which gives $S_1^{t}(X)$, and multiplied with $S_2(X)$ to obtain an  $S^{m,n}$. Then, by applying softmax operation, gives ab attention map$S_a$ for pixel value (m,n) of the shape $HW\times HW$ . is given as:


\begin{equation}
S_a^{m,n} = \frac{\exp(S^{m,n})}{\sum_{m=1}^{H \times W} \exp(S^{m,n})}
\end{equation}
The value $S^{m,n}_a$ represents how much the $m_{th}$ pixel influences the $n_{th}$ pixel in the feature map. Finally, $S^{mn}_a$ is multiplied with $S_3(X)$ and reshaped the output as $\in \mathbb{R}^{C \times H \times W }$. The final spatial attention with learnable parameter $\lambda$ and initial feature map $F_n $ is given as follows:
\begin{equation}
S_n^{final} = \lambda \sum_{m=1}^{HW} (S_a^{mn} \cdot S_3(X)) + F_n
\end{equation}
Similarly for channel attention an input feature map $F \in \mathbb{R}^{C \times H \times W}$ is given as 
\begin{equation}
C_a^{m,n} = \frac{\exp(C^{m,n})}{\sum_{m=1}^{H \times W} \exp(C^{m,n})}
\end{equation}
where $C^{mn}_a$ denotes the influence of the $m$-th channel on the $n$-th channel. The final feature map, $C_{\text{final}}$, with a size of $C \times H \times W$, is computed as:
\begin{equation}
C_n^{final} =  \mu \sum_{m=1}^{HW} C_a^{mn} \cdot C_3(X) + C_n
\end{equation}
where $\mu$ is a learnable parameter, and $C_n$ is the initial feature map.
\subsection{Loss Function}
We adopt the Euclidean distance to measure the pixel-wise difference between estimated density maps and their corresponding ground truth as in previous works\cite{li2018csrnet}\cite{zhang2016single}. This approach allows the network to focus on individual pixels and effectively learn the density distribution of the crowd. Thus, the regression loss is given as.
\begin{equation}
L(\Theta) = \frac{1}{2N} \sum_{i=1}^{N} \left( \text{D}(I_i; \Theta) - \text{GT}(I_i) \right)^2
\end{equation}
Where N is the size of the training set, and $I_i$ represents the $i_{th}$ input image.$\text{D}(I_i; \Theta)$ is the output and $\text{GT}(I_i)$ is ground truth density map of image. $L(\Theta)$  denotes the loss between the ground-truth density map and the estimated density map.\\

\section{Experiments}
\textbf{Implementation Details: } We integrate three layers of FGA module to baseline method CSRNet\cite{li2018csrnet} and CANNet\cite{liu2019context} just after VGG-16 feature extractor and train the network end-to-end. We use the Adam optimizer\cite{kingma2017adam} with a learning rate of 1e-5. Momentum is between 0.93 and 0.99 for oscillation damping and a weight decay rate 0.001. The training comprises 100 epochs to optimize performance and improve generalization.\\
\textbf{Evaluation Metrics:} We use the mean absolute error (MAE) and root mean square error (MSE) as evaluation metrics for crowd density estimates. It is defined as $\text{MAE} = \frac{1}{N}\sum_{i=1}^{n} |y_i - \hat{y}_i|$ and \textbf{MSE} is $ = \sqrt{\frac{1}{N}\sum_{i=1}^{n} (y_i - \hat{y}_i)^2}$. where N is the number of the test images, $y_i$ and $\hat{y}_i$ are the ground truth and estimated counts of the $i_{th}$ image, respectively.\\
\textbf{Ground Truth Generation:} 
We employ geometry-adaptive kernels to handle highly congested scenes in line with the density map generation method described in \cite{zhang2016single}. By applying a normalized Gaussian kernel to blur each head annotation, we generate ground truth that considers the spatial distribution of all images in each dataset. The geometry-adaptive kernel is defined as follows:
\begin{equation}
F(x) = \sum_{i=1}^{N} \delta(x - x_i) \times G_{\sigma_i}(x)
\end{equation}
where \(\sigma_i = \beta d_i\) for each targeted object \(x_i\) in the ground truth \(\delta\), with \(d_i\) representing the average distance of \(k\) nearest neighbors. The density map is generated by convolving \(\delta(x - x_i)\) with a Gaussian kernel having a parameter \(\sigma_i\) (standard deviation), where \(x\) denotes the pixel position in the image. In our experiments, we adhere to the configuration in \cite{zhang2016single} with \(\beta = 0.3\) and \(k = 3\). In sparse crowds, we adapt the Gaussian kernel to the average head size to blur all annotations.

\begin{figure*}[ht]
\centering

\begin{minipage}{0.15\textwidth}
  \centering
  \textcolor{black}{\textbf{a)}} \\
\end{minipage}%
\hfill
\begin{minipage}{0.8\textwidth}
  \centering
  \includegraphics[width=\textwidth]{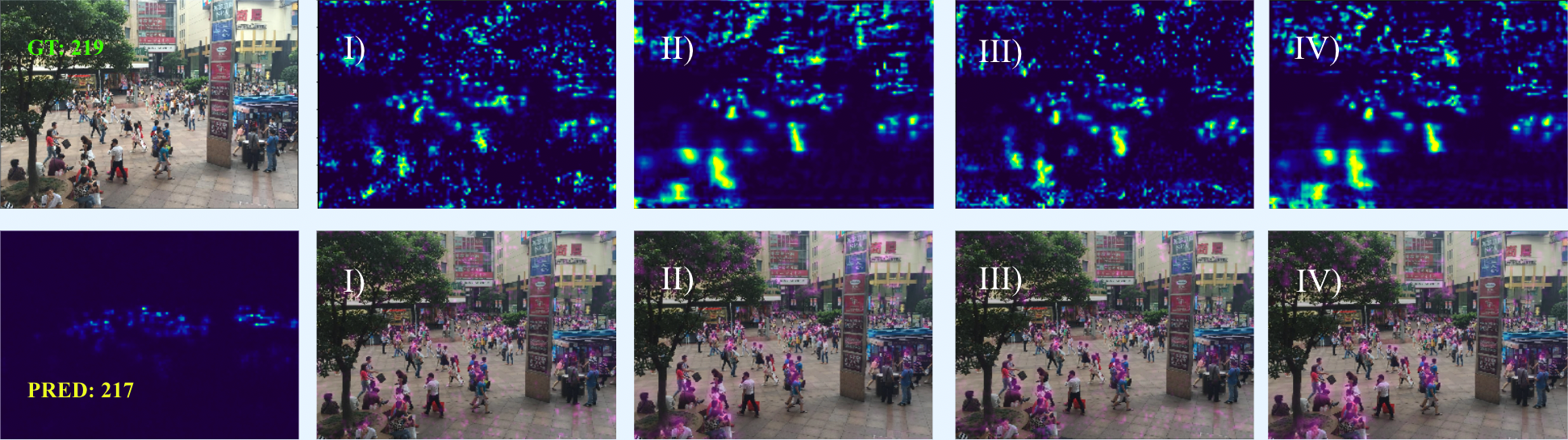}
\end{minipage}

\vskip 0.5\baselineskip

\begin{minipage}{0.15\textwidth}
  \centering
  \textcolor{black}{\textbf{b)}} \\
\end{minipage}%
\hfill
\begin{minipage}{0.8\textwidth}
  \centering
  \includegraphics[width=\textwidth]{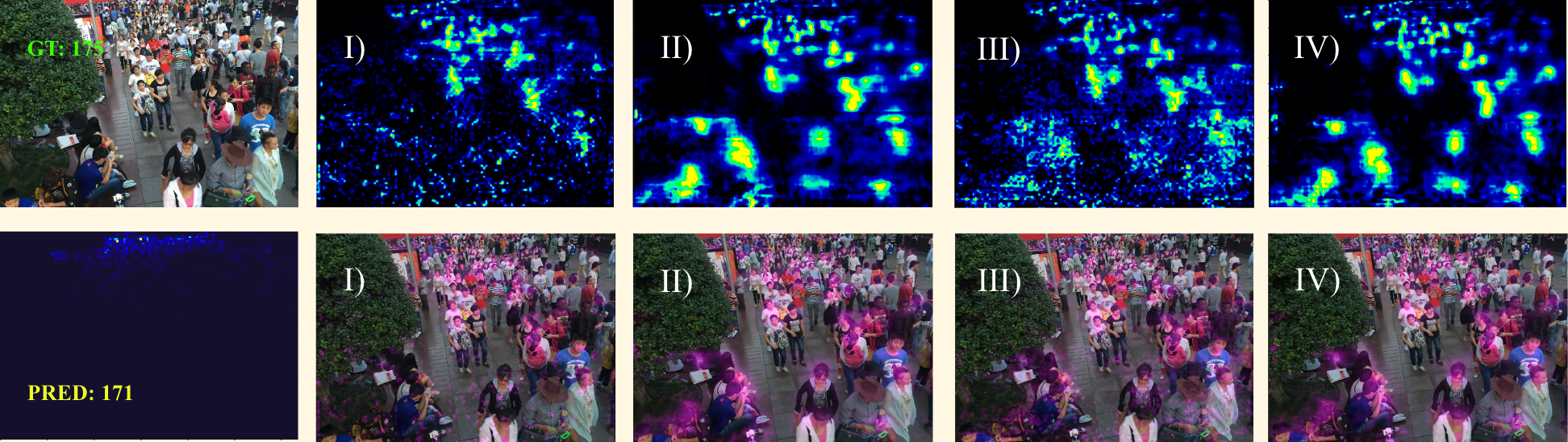}
\end{minipage}

\vskip 0.5\baselineskip

\begin{minipage}{0.15\textwidth}
  \centering
  \textcolor{black}{\textbf{c)}} \\
\end{minipage}%
\hfill
\begin{minipage}{0.8\textwidth}
  \centering
  \includegraphics[width=\textwidth]{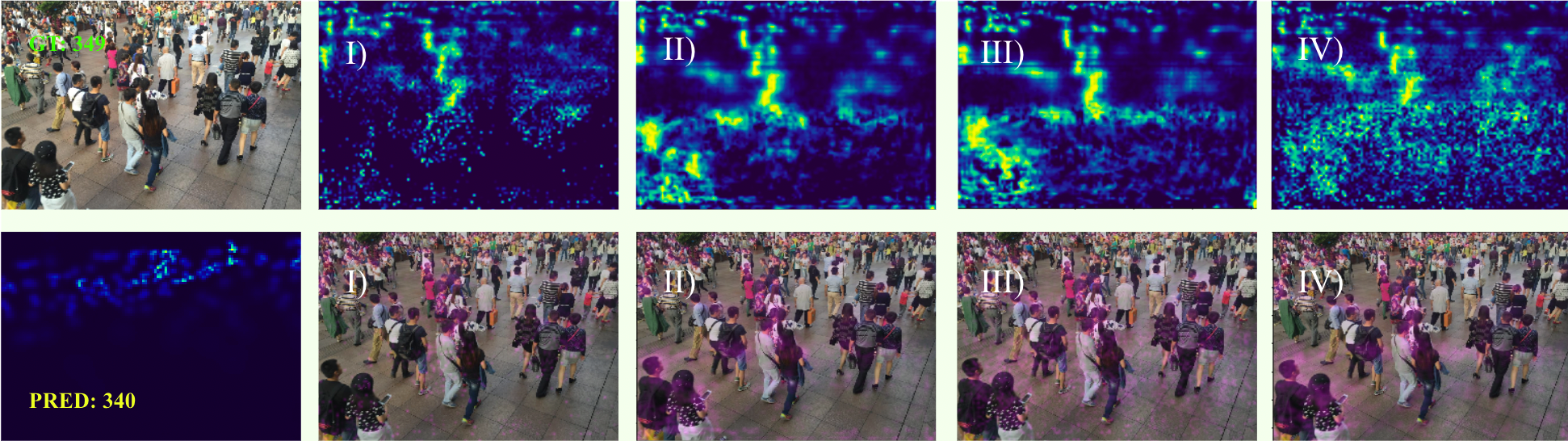}
\end{minipage}

\caption{\textbf{Gradient activation maps of final layer for different models with and without FGA module: } \textcolor{black}{\textbf{a)}},  \textcolor{black}{\textbf{b)}} and \textcolor{black}{\textbf{c)}} are three sample images with different degrees of crowd dispersion, representing high, moderate, and low levels of crowd dispersion, respectively. Each sample image is analysed using gradient activation maps to understand area of activation around a head. I), II), III), IV) represents the performance of FGA + CSRNet, CSRNet, FGA + CANNet and CANNet on the same sample image when provided as input. Each gradient activation map has its reference image fused with activations below.}
\label{fig:grad_maps}
\end{figure*}

\subsection{ShanghaiTech Dataset:-}
\begin{table}[h]
    \centering
    \resizebox{0.4\textwidth}{!}{
        \begin{tabular}{l|ll|ll}
            \hline
            \textbf{Models} & \multicolumn{2}{c|}{\textbf{SHA}} & \multicolumn{2}{c}{\textbf{SHB}} \\
            & \textbf{MSE} & \textbf{MAE} & \textbf{MSE} & \textbf{MAE} \\ \hline
            Bayesian+\cite{ma2019bayesian} & 101.8 & 62.8 & 12.7 & 7.7 \\
            DMCNet\cite{wang2023dynamic} & 84.5 & 58.4 & 13.6 & 8.6 \\
            P2PNet \cite{song2021rethinking} & 85.0 & 52.7 & 9.9 & 6.2 \\
            GAUNet\cite{cheng2022rethinking} & 89.1 & 54.8 & 9.9 & 6.2 \\
            NASCount\cite{hu2020count} & 93.40 & 56.7 & 102.0 & 6.7 \\
            DMCount\cite{wang2020distribution} & 95.7 & 59.7 & 11.8 & 7.4 \\
            BBANet\cite{hou2020bba} & 93.8 & 63.8 & 12.0 & 7.8 \\
            ASNet\cite{jiang2020attention} & 90.1 & 57.7 & - & - \\
            DensityToken\cite{hu2023densitytoken} & 100.5 & 64.4 & 17.1 & 9.20 \\
            MAN\cite{lin2022boosting} & 90.3 & 56.8 & - & - \\
            IF-CKT+\cite{cai2023explicit} & 16.8 & 10.9 & - & - \\
            DA2Net\cite{zhai20232} & 128.4 & 74.1 & 13.2 & 7.9 \\
            CANNet\cite{liu2019context} & 100.0 & 62.3 & 12.2 & 7.8 \\
            CSRNet\cite{li2018csrnet} & 115.0 & 68.2 & 16.0 & 10.6 \\
            \textbf{CANNet + FGA} & \textbf{90.6} & \textbf{61.1} & \textbf{10.4} & \textbf{6.9} \\
            \textbf{CSRNet + FGA} & \textbf{110.4} & \textbf{66.8} & \textbf{13.5} & \textbf{8.4} \\
            \hline
        \end{tabular}
    }
    \caption{Evaluations on ShanghaiTech Datasets}
\end{table}
ShanghaiTech\cite{zhang2016single} dataset is a popular benchmark for evaluating crowd-counting algorithms due to its crowd variations. The ShanghaiTech dataset, crucial for crowd-counting research, comprises 1,198 high-resolution images from the ShanghaiTech University campus. Part A has 482 images; Part B has 716, totalling around 330,000 annotated individuals. The FGA module improves both CSRNet\cite{li2018csrnet} and CANNet\cite{liu2019context} over ShanghaiTech Datasets with an 18.5\% and 20.7\% improvement on MSE and MAE on ShanghaiTech-B dataset for CSRNet + FGA and 14\% and 11\% for MSE and MAE metrics for CANNet + FGA model.

\subsection{UCF-CC-50 Dataset:-}
\begin{table}[h]
    \centering
    \resizebox{0.37\textwidth}{!}{
        \begin{tabular}{l|ll}
            \hline
            \textbf{Models} & \textbf{MSE} & \textbf{MAE} \\ \hline
            Bayesian+\cite{ma2019bayesian} & 308.2 & 229.3 \\
            P2PNet\cite{song2021rethinking} & 256.1 & 172.7 \\
            GAUNet\cite{cheng2022rethinking} & 256.5 & 186.3 \\
            NASCount\cite{hu2020count} & 297.3 & 208.4 \\
            DMCount\cite{wang2020distribution} & 291.5 & 211.0 \\
            BBANet\cite{hou2020bba} & 316.9 & 230.5 \\
            ASNet\cite{jiang2020attention} & 251.6 & 174.8 \\
            DA2Net\cite{zhai20232} & 237.0 & 167.0 \\
            CANNet\cite{liu2019context} & 243.7 & 212.2 \\
            CSRNet\cite{li2018csrnet} & 397.5 & 266.1 \\
            \textbf{CANNet + FGA} & \textbf{223.0} & \textbf{197.2} \\
            \textbf{CSRNet + FGA} & \textbf{378.2} & \textbf{252.4} \\
            \hline
        \end{tabular}
    }
    \caption{Evaluations on UCF-CC-50 Dataset}
\end{table}
The UCF-CC-50\cite{6619173} dataset consists of 50 high-resolution images, meticulously annotated to provide accurate head counts. It serves as a benchmark for crowd-counting algorithms, encompassing diverse environments and presenting challenges in varying crowd densities. The application of the FGA module shows an improvement in the performance of the CSRNet\cite{li2018csrnet} and CANNet\cite{liu2019context} baselines, as shown in Table II.

\subsection{JHU++ crowd Dataset:-}
We also evaluate our Proposed Approach on the JHU++ crowd dataset\cite{sindagi2019pushing}, which has 4,372 images with a total of 1.51 million annotations of varying crowd orientations. Table III. The FGA module can significantly reduce the baseline MSE and MAE metrics, with CANNet\cite{liu2019context} improving by 22.8 and 7.9 points, respectively. CSRNet\cite{li2018csrnet} improved by 25.5 and 16.7 points on MSE and MAE metrics.
\begin{table}[h]
    \centering
    \resizebox{0.37\textwidth}{!}{
        \begin{tabular}{l|ll}
            \hline
            \textbf{Models} & \textbf{MSE} & \textbf{MAE} \\ \hline
            DMCNet\cite{wang2023dynamic} & 246.9 & 69.6 \\
            GAUNet \cite{cheng2022rethinking}& 245.1 & 58.2 \\
            DMCount\cite{wang2020distribution} & 261.4 & 66.0 \\
            MAN\cite{lin2022boosting} & 209.9 & 53.4 \\
            DA2Net & 204.3 & 111.7 \\
            CANNet\cite{liu2019context} & 278.4 & 63.2 \\
            CSRNet\cite{li2018csrnet} & 309.2 & 85.9 \\
            \textbf{CANNet + FGA }& \textbf{255.6} & \textbf{55.3} \\
            \textbf{CSRNet + FGA} & \textbf{283.7} & \textbf{69.2} \\
            \hline
        \end{tabular}
    }
    \caption{Evaluations on JHU++ Dataset}
\end{table}

\subsection{Grad-CAM analysis on crowds having different degrees of dispersion in an image}
We explore gradient activation maps using Grad-CAM\cite{selvaraju2017grad} as shown in Figure 6 to understand and break down the impact of the FGA module on CSRNet\cite{li2018csrnet} and CANNet\cite{liu2019context} when presented with crowds having low, moderate and high amounts of dispersion. We find that the FGA module sharpens the area of activation in all cases with noticeable improvement in high and moderate levels of dispersion, showing the efficacy of the FGA module in existing crowd estimation networks and validating utility of FGA modules.
\subsection{Ablations:-}\textbf{Effectiveness of Attention:-} To verify the impact of the individual parts of the attention module, we perform a series of experiments on the ShanghiTech-B dataset. Our analysis investigates the contributions of Fast Fourier Convolutions(FFC), spatial attention (SA) and Channel Attention(CA) to our architecture. We comprehend and assess the relative importance of these two components within the overall framework. Table IV shows that the combined effect of FFC, Channel attention, and Spatial attention reduces the MSE and MAE metrics, whereas partial implementation partially degrades the baseline performance. We also see that Individual use of FFC is not efficient in the context of crowd counting. 
\begin{table}[h]
\centering
\begin{tabular}{l|ll|ll}
\hline
\textbf{Combination} & \multicolumn{2}{c|}{\textbf{CANNet}} & \multicolumn{2}{c}{\textbf{CSRNet}} \\ \cline{2-5} 
& \textbf{MAE} & \textbf{MSE} & \textbf{MAE} & \textbf{MSE} \\ \hline
FFC & 14.1 & 21.8 & 18.7 & 27.3 \\ 
FFC + SA & 9.9 & 12.3 & 12.2 & 18.8 \\ 
FFC + CA & 8.7 & 12.5 & 11.9 & 18.0 \\ 
FFC + CA + SA & 6.9 & 10.4 & 8.4 & 13.5 \\ \hline
\end{tabular}
\caption{Ablation study over CSRNet\cite{li2018csrnet} and CANNet\cite{liu2019context} with varying internal configuration}
\end{table}
\\
\subsection{Counting on varying crowd densities:-} We analyse counting samples with different crowd densities in the FGA + CANNet and FGA + CSRNet configurations, as shown in Figure 5. On closer inspection, we found that our model performs exceptionally well in low-density, moderate-density, and high-density crowds. We see a minor decrease in performance in high-density crowds. This needs further analysis to explain why Fourier-guided attention slightly decreases in high crowd densities and has been left as a future work in this domain.
\section{Conclusion}
In this paper, we propose Fourier-guided attention using attention-assisted convolutions in the frequency domain and real domain for crowd count estimation, which leverages the effective combination of full-scale global and local feature extraction in a single unit.  It could be integrated into CNN-based crowd count estimation methods to improve performance. Experiments conducted across four benchmark datasets with two baseline models demonstrate that it can significantly improve the baselines. To check individual robustness and relative performance to state-of-the-art methods, we perform a benchmark analysis and a qualitative analysis along with an ablation study of individual components of the model.

\bibliographystyle{IEEEtran}
\bibliography{main}

\end{document}